\documentclass{llncs}

\usepackage[table]{xcolor}
\usepackage{array}
\usepackage{multirow}
\usepackage{makecell}
\usepackage{booktabs}
 \usepackage[plainpages=false, colorlinks=true,
   citecolor=blue, filecolor=black, linkcolor=blue,
   urlcolor=blue]{hyperref}
\usepackage{colortbl}
\usepackage{subcaption}

\usepackage{amsmath,amssymb,amsbsy,amsfonts}
\usepackage{algorithm,algorithmic}
\usepackage{pdflscape}
\usepackage{multirow,multirow}

\usepackage{bbold}

\usepackage{siunitx}
\usepackage[pdftex]{graphicx}
\usepackage{graphics,epsfig,float,color}
\graphicspath{{figs}}
\usepackage{wrapfig}
\usepackage[sort,compress]{cite}

\graphicspath{{figures/}{tables/}}

\definecolor{light-gray}{gray}{0.80}
\definecolor{ColumnGray}{gray}{0.95}
\newcolumntype{g}{>{\columncolor{ColumnGray}}c}
\newcolumntype{d}{>{\columncolor{gray!40}}l}

\newcommand\Ga{ \rowcolor{gray!0}}

\newcommand\Gd{ \rowcolor{gray!45}}

\graphicspath{{figures/}}
\DeclareGraphicsExtensions{.pdf,.png}

\title{A Novel Domain Adaptation Framework for Medical Image Segmentation}
\titlerunning{}  

\begin{document}
\pagestyle{headings}  

\author{
\small
  Amir Gholami\inst{1}\thanks{Authors contributed equally} \and Shashank Subramanian\inst{2}$^*$  \and Varun Shenoy\inst{1} \and Naveen Himthani\inst{2}\and \\ Xiangyu Yue \inst{1}\and
  Sicheng Zhao \inst{1} \and Peter Jin \inst{1} \and George Biros\inst{2} \and Kurt Keutzer\inst{1}
}
\authorrunning{Gholami et al.} 
\institute{
  University of California Berkeley
\and
The University of Texas at Austin
}

\maketitle

\begin{abstract}
We propose a segmentation framework that uses deep neural networks and introduce two innovations. First, we describe a biophysics-based domain adaptation method. Second, we propose an automatic method to segment white and gray matter, and cerebrospinal fluid, in addition to tumorous tissue.
Regarding our first innovation, we use a domain adaptation framework that combines a novel multispecies biophysical tumor growth model with a generative adversarial model to create realistic looking synthetic multimodal MR images with known segmentation. 
Regarding our second innovation, we propose an automatic approach to enrich available segmentation data by computing the segmentation for healthy tissues. This segmentation, which is done using diffeomorphic image registration between the BraTS training data and a set of prelabeled atlases, provides more information for training and reduces the class imbalance problem. 
Our overall approach is not specific to any particular neural network and can be used in conjunction with existing solutions.  We demonstrate the performance improvement using a 2D U-Net for the BraTS'18 segmentation challenge.
Our biophysics based domain adaptation achieves better results, as compared to the existing state-of-the-art GAN model
used to create synthetic data for training.
\end{abstract}

 \label{s:abstract}

\section{Introduction} \label{s:intro}

Automatic segmentation methods have the potential to provide accurate and reproducible labels
leading to improved tumor prognosis and treatment planning, especially for cases where access to expert radiologists is limited.

In the BraTS competition, we seek to segment multimodal MR images of glioma patients. 
Common brain MRI modalities include  post-Gadolinium T1 (used to enhance contrast and  visualization of the blood-brain barrier), T2 and FLAIR (to
highlight different tissue  fluid intensities), and T1. We use the data for these four modalities to generate the segmentations using a methodology that we outline below.

\textit{Contributions:} In most image classification tasks deep neural 
networks (DNNs) have been a very powerful technique that tends to 
outperform other approaches and BraTS is no different. From past BraTS 
competitions two main DNN architectures have emerged: DeepMedic~\cite{ISLES2015}
and U-Net~\cite{Ronneberger15}.
How can we further improve this approach? Most research efforts have been on further improving these architectures,
as well as coupling them with post-processing and ensemble techniques.
In our work here, we propose a framework to work around the relatively small training datasets used in the BraTS competition. 
Indeed, in comparison to other popular classification challenges like ImageNet~\cite{deng2009imagenet} (which consists of one million images for training),
the BraTS training set contains only 285 instances (multimodal 3D MR images), a number that is several orders of magnitude smaller than the typical number of instances required for DNNs to work well. These observations have motivated this work, whose contributions we summarize below.
\begin{figure}[!htbp]
\centering
\begin{subfigure}{.33\textwidth}
    \includegraphics[height=\linewidth, width=\linewidth]{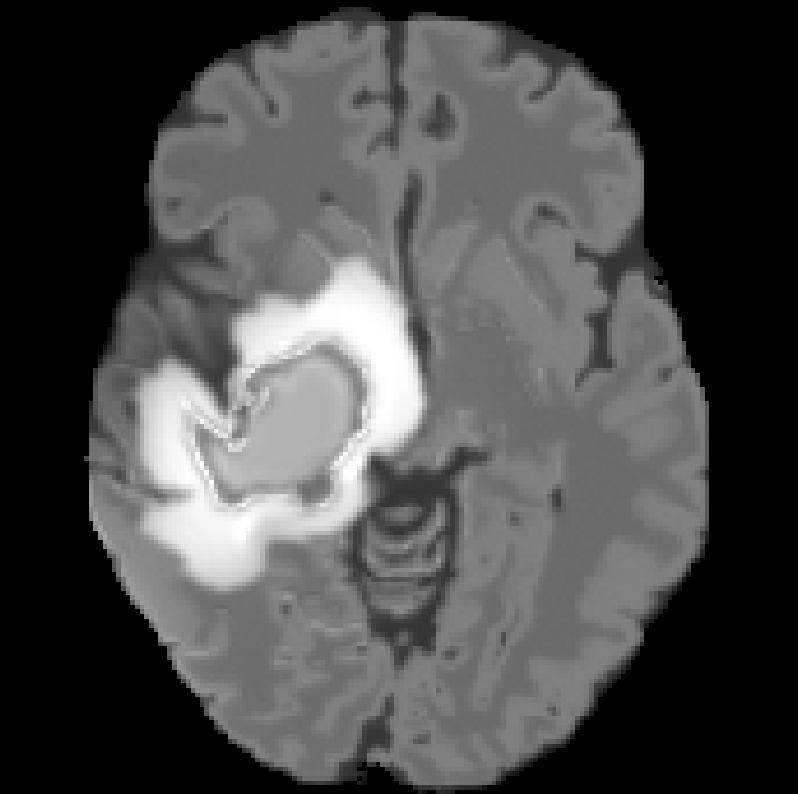}
    \caption{}
\end{subfigure}%
\begin{subfigure}{.33\textwidth}
    \includegraphics[height=\linewidth, width=\linewidth]{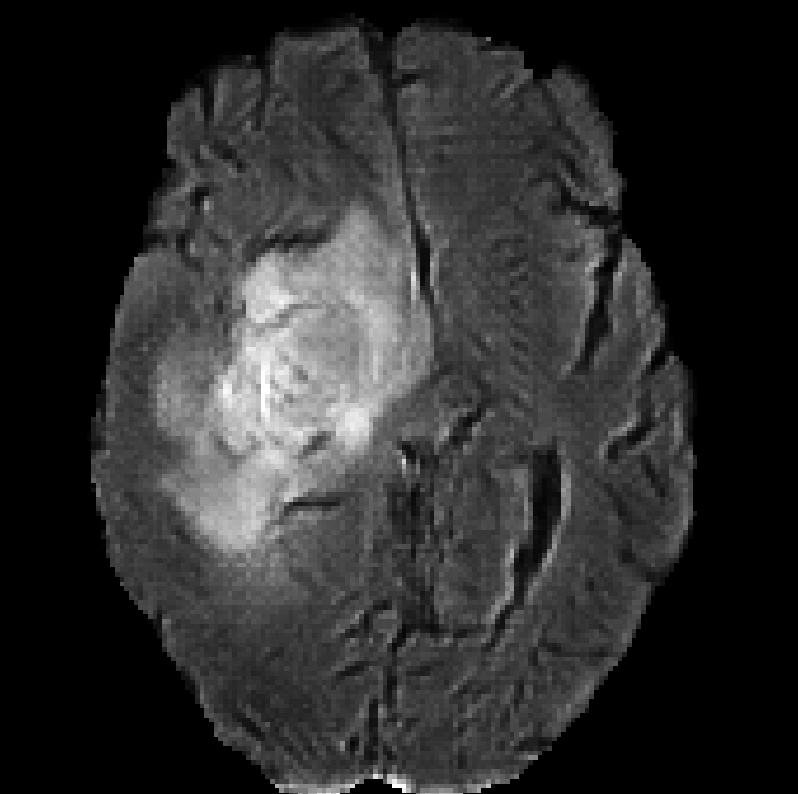}
    \caption{}
\end{subfigure}%
\begin{subfigure}{.33\textwidth}
    \includegraphics[height=\linewidth, width=\linewidth]{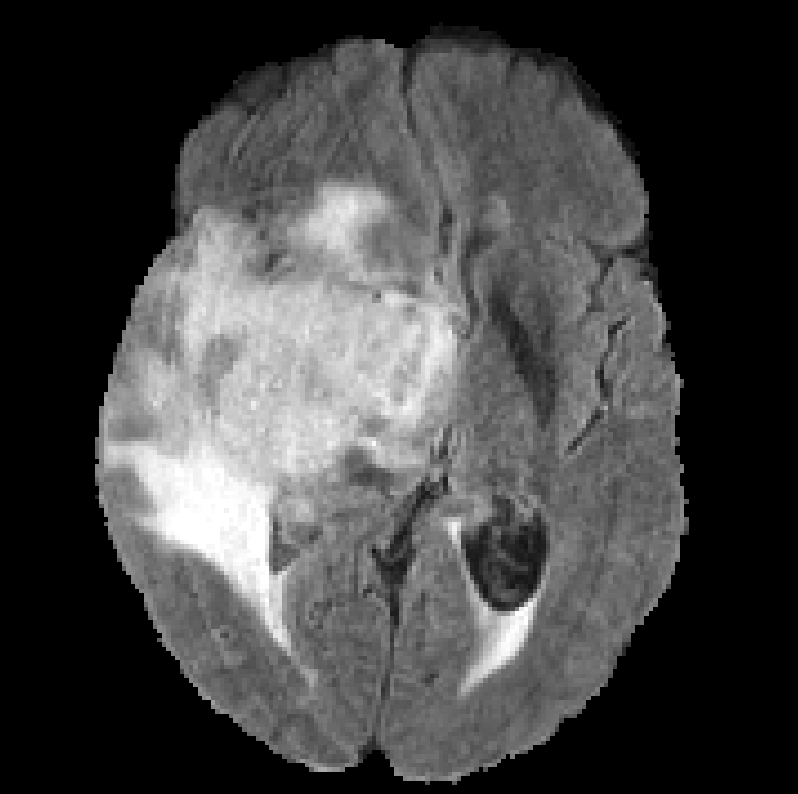}
    \caption{}
\end{subfigure}
\caption{Domain adaptation results: (a) represents a synthetic FLAIR brain image, (b) represents the domain adapted synthetic FLAIR image, (c) represents the real BraTS FLAIR image. As we can see from the intensity distributions, the values in the adapted images are qualitatively closer to the real images.}
\label{f:domain_adaptation}
\end{figure}

\begin{enumerate}
    \item {\bf Data augmentation:} We propose a biophysics based domain adaptation strategy to add synthetic tumor-bearing MR images to the training examples. There have been many notable works to simulate tumor growth (see ~\cite{oden2016toward,ivkovic-swanson-rosenfeld-e12,HawkinsDaarud:2013a,hawkins2012numerical,lima2016selection,gholami_thesis}). We use an in-house PDE based multispecies tumor growth model~\cite{masseffect} to simulate synthetic tumors. Since simulated data does not contain the correct intensities distribution of a real MR image, we train an auxiliary neural network to transform the simulated images to match real MRIs. This network gets a multimodal input and transforms this data to match the distribution of BraTS images by imposing certain cycle consistency constraints. As we will show, this is a very promising approach.
    
    \item {\bf Extended segmentation:} We extend the segmentation to the healthy parenchyma. This is done in two steps. First, we segment the training dataset using an atlas-based ensemble registration (using an in-house diffeomorphic registration code). Second, we train our DNN network to segment both tumor and healthy tissue (four classes, glial matter, cerebrospinal fluid, gray, and white matter).
    Our approach adds important information about healthy tissue delineation, which is actually used by radiologists.
    It also reduces the inherent class imbalance problem. 
\end{enumerate}

Our data augmentation is different than the recent work of~\cite{shin2018medical}, which uses GANs to automatically generate data.
To the best of our knowledge, our work here is the first to use a biophysics based  domain adaptation framework for automatic data generation, and
our approach achieves five percentage points higher dice score as compared to~\cite{shin2018medical}, even though we use a 2D neural
network architecture (which has suboptimal performance as compared to 3D network used in~\cite{shin2018medical}).

\textit{Related work:} Recently, deep learning approaches using convolutional neural networks (CNNs) have demonstrated excellent performance in 
semantic segmentation tasks in medical imaging.  Seminal works for segmentation stem from fully-convolutional networks (FCNs)~\cite{Long14}. 
U-Net~\cite{Ronneberger15} is another popular architecture for medical segmentation, which merges feature maps from the contracting path of an FCN
to its expanding path to preserve local contextual information. Multiscale information is often incorporated by using parallel convolutional 
pathways of various resolutions~\cite{KAMNITSAS2017} or by using dilated convolutions and cascading network architectures~\cite{Wang2017}. 
Post-processing and ensemble methods are also usually used after training with these models. The most commonly used post processing step is Conditional 
Random Fields (CRF)~\cite{KAMNITSAS2017}, which
has been found to significantly reduce false positives and sharpen the segmentation.
Ensembling is also very important to reduce overfitting with deep neural networks. The winning algorithm of the Multimodal \underline{Bra}in \underline{T}umor Image \underline{S}egmentation Benchmark (BraTS) challenge in 2017 was based on Ensembles of Multiple Models and Architectures (EMMA)~\cite{EMMA2017}, which bagged a heterogeneous collection of networks (including DeepMedic (winner of ISLES 2015~\cite{ISLES2015}), U-Nets and FCNs) to build a robust and generic segmentation model.

There are established techniques to address training with small datasets, such as regularization, or ensembling, which was the approach taken by the winning team of BraTS'17. However, in this paper we propose an orthogonal method to address this problem.

\textit{Limitations: } Currently, our framework only supports 2D domain transformations. Hence, we are limited to transforming 3D brains slice-by-slice and using only 2D neural network architectures. This is sub-optimal as 3D CNNs can demonstrably utilize volumetric medical imaging data more efficiently leading to better and more robust performance (see~\cite{KAMNITSAS2017, EMMA2017, isensee2017}). Hence, extending our framework to 3D is the focus of our future work and can potentially lead to greater improvements in performance.

The outline of the paper is as follows. In \S\ref{s:methods}, we discuss the methodology for domain adaptation (\S\ref{s:domain_adaptation}), and the whole brain segmentation (\S\ref{s:registration}) . In \S\ref{s:results} we present preliminary results for the BraTS'18 challenge~\cite{ISLES2015,bakas-davatzikos-e17,bakas-davatzikos-e17b,bakas-davatzikos-e17c}. Our method achieves a Dice score of [79.15,90.81,81.91] for enhancing tumor, whole tumor and tumor core, respectively for the BraTS'18 validation dataset.

\section{Methods} \label{s:methods}

\subsection{Domain Adaptation} \label{s:domain_adaptation}

As mentioned above, one of the main challenges in medical imaging is the
scarcity of training data. To address this issue,
we use a novel domain adaptation strategy and generate synthetic tumor-bearing 
MR images to enrich the training dataset. 
This is performed by first solving an in-house PDE based
multispecies tumor model using an atlas brain~\cite{masseffect}.
This model captures the time evolution of enhancing and necrotic tumor concentrations 
along with tumor-induced brain edema. The governing equations for the model are 
reaction-diffusion-advection equations for the tumor species along with a diffusion 
equation for oxygen and other nutrients. We couple this model with linear elasticity 
equations with variable elasticity material properties to simulate the deformation of 
surrounding brain tissue due to tumor growth, also known as ``mass effect". 
However, this data cannot be used directly due to the difference in intensity 
distributions between a BraTS MRI scan and a synthetic MRI scan. Directly using synthetic
data during the training process will adversely guide the neural network to learn 
features which do not exist in a real MR image, resulting in poor performance.

To address this issue, we use CycleGAN~\cite{CycleGAN2017} to perform domain adaptation 
from the generated synthetic data to the real BraTS images. This is done by learning a 
mapping $G : X \rightarrow Y$ such that the distribution of images from $G(X)$ is 
indistinguishable from the distribution $Y$ using an adversarial loss, as shown in 
Fig.~\ref{f:cyclegan}. Here, $X$ is the simulated tumor data, and $Y$ is the 
corresponding data which matches the BraTS distribution. Because this mapping is highly 
under-constrained, it is coupled with an inverse mapping $F: Y\rightarrow X$ and a cycle 
consistency loss is introduced to enforce $F(G(X)) \approx X$ (and vice versa).

\begin{figure}[!htbp]
\centering
\includegraphics[width=0.3\textwidth]{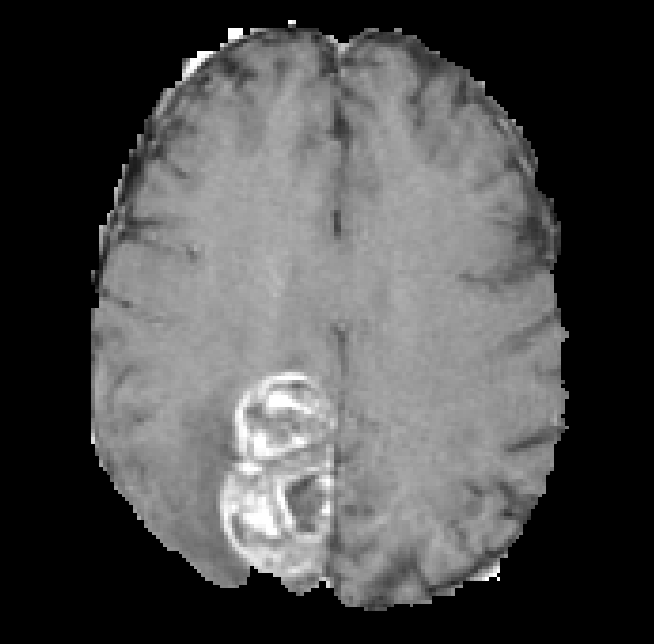}
\includegraphics[width=0.3\textwidth]{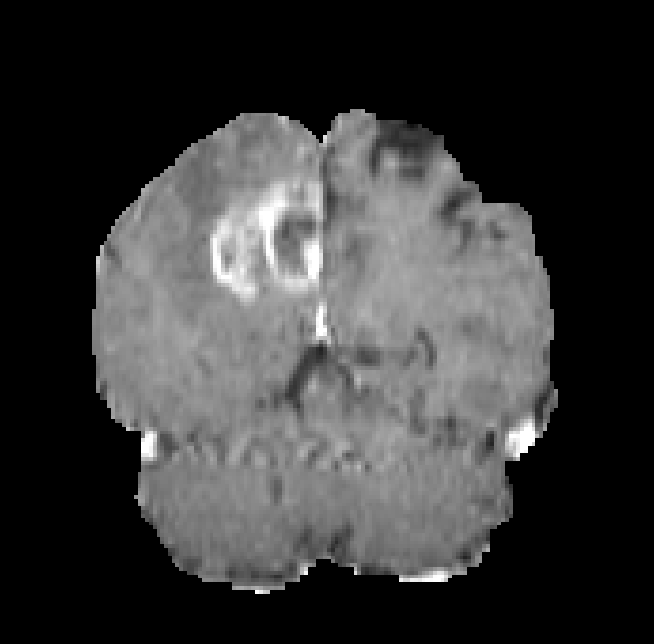}
\includegraphics[width=0.3\textwidth]{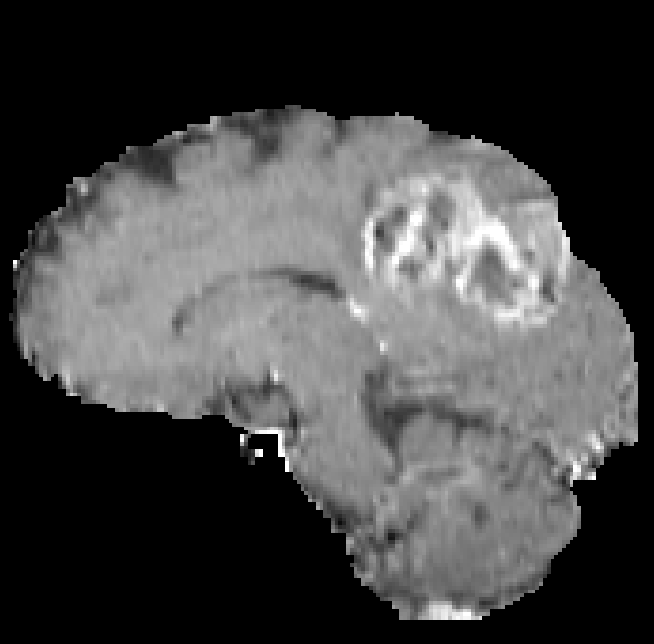}\\
\includegraphics[width=0.3\textwidth]{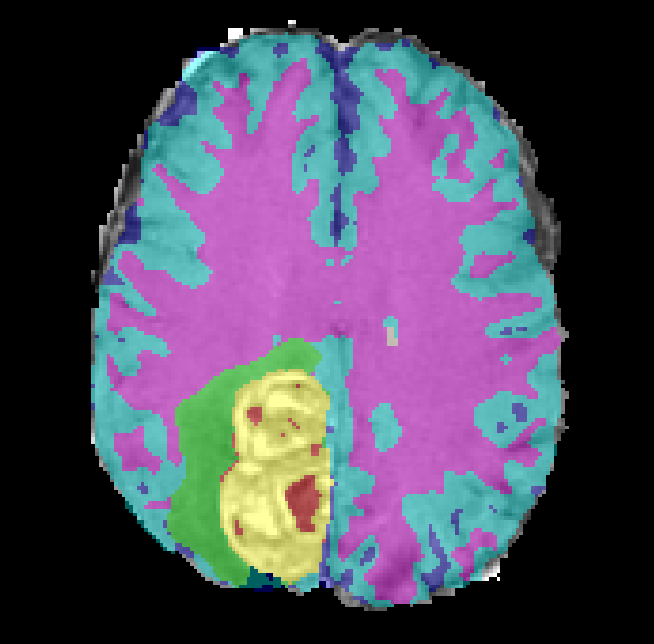}
\includegraphics[width=0.3\textwidth]{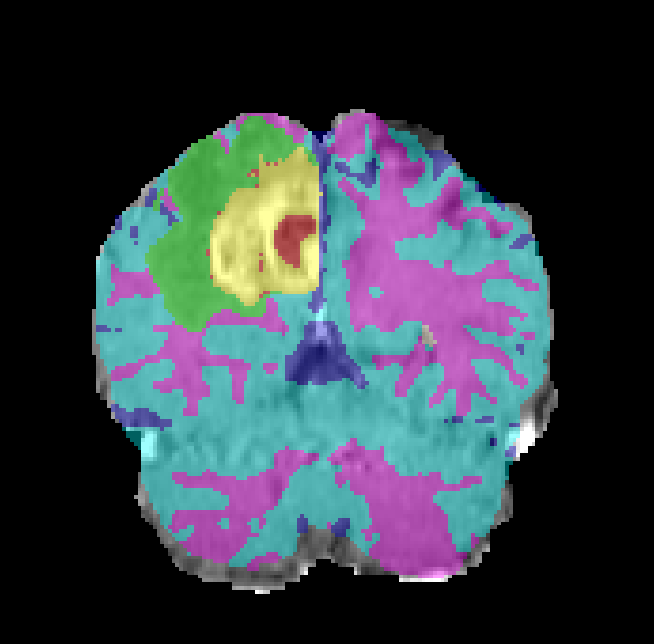}
\includegraphics[width=0.3\textwidth]{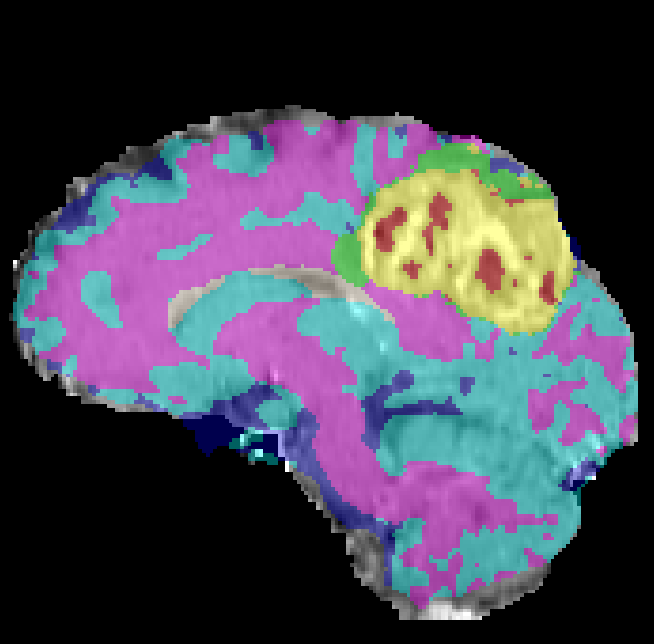}
\caption{
  (\textit{Top row}): The original T1ce image for Brats18\_TCIA02\_135\_1 training data is shown for different views (axial, coronal and sagittal).
  (\textit{Bottom row}): The corresponding extended segmentation for healthy cells computed by solving a 3D registration problem with a
  segmented atlas. We overlay the BraTS tumor segmentation with the registered segmentation to get the final results.
}
\label{f:registration}
\end{figure}

For training the domain adaptation network, we first computationally simulate synthetic 
tumors in a healthy brain atlas, located approximately at the whole tumor center 
taken from each BraTS image. Hence, every synthetic tumorous brain is paired with 
the corresponding data from a real BraTS image. Then, we perform a pre-processing step to
transform our synthetic results to intensities. We produce a segmentation map for every 
tissue (healthy and tumorous) class and sample intensities for each class from a real MRI
scan. We assign these sampled intensities to every voxel in our synthetic segmentation 
map to finally obtain our synthetic MRI scans.
Then, we train with these pre-processed synthetic MRI scans and their corresponding BraTS
images. Samples of our adaptation results are shown Fig.~\ref{f:domain_adaptation}, which
demonstrate an almost indistinguishable adaptation of the simulated data with the real images.

\subsection{Whole Brain Segmentation With Healthy Tissues} \label{s:registration}

\begin{figure}[!htbp]
\centering
\includegraphics[width=0.3\textwidth]{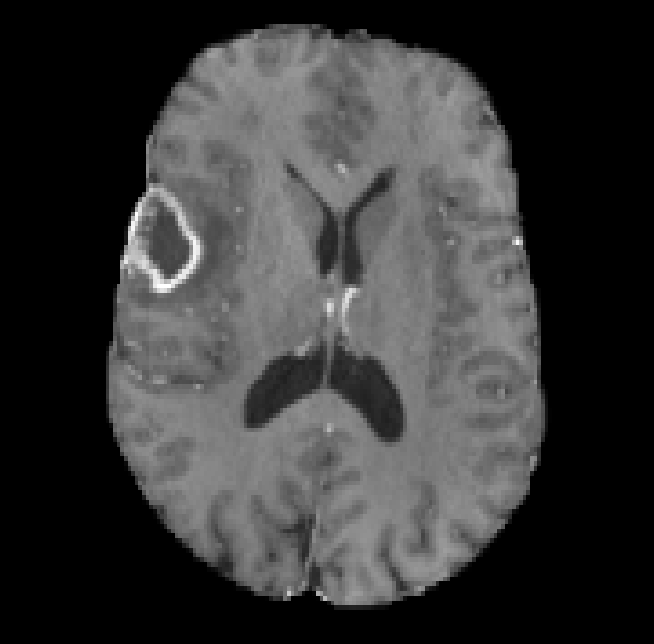}
\includegraphics[width=0.3\textwidth]{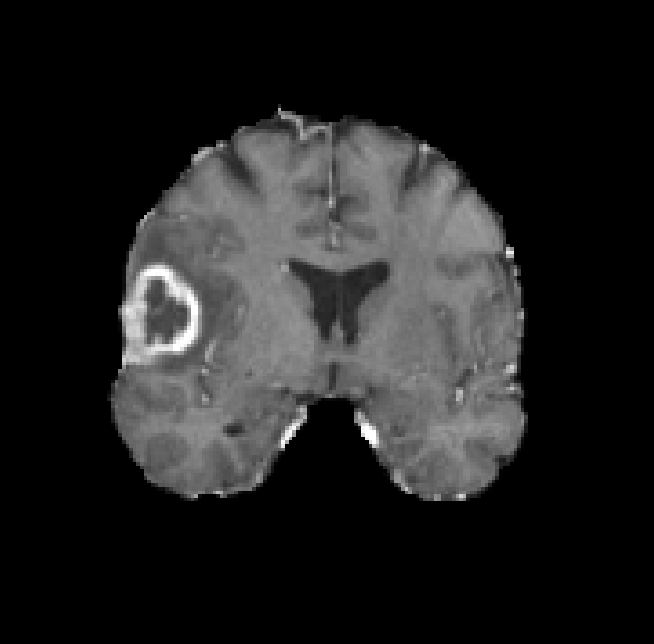}
\includegraphics[width=0.3\textwidth]{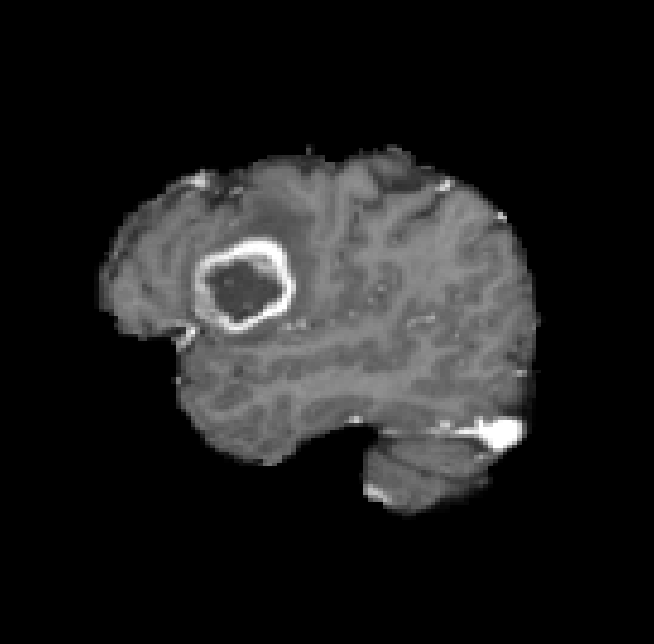}
\includegraphics[width=0.3\textwidth]{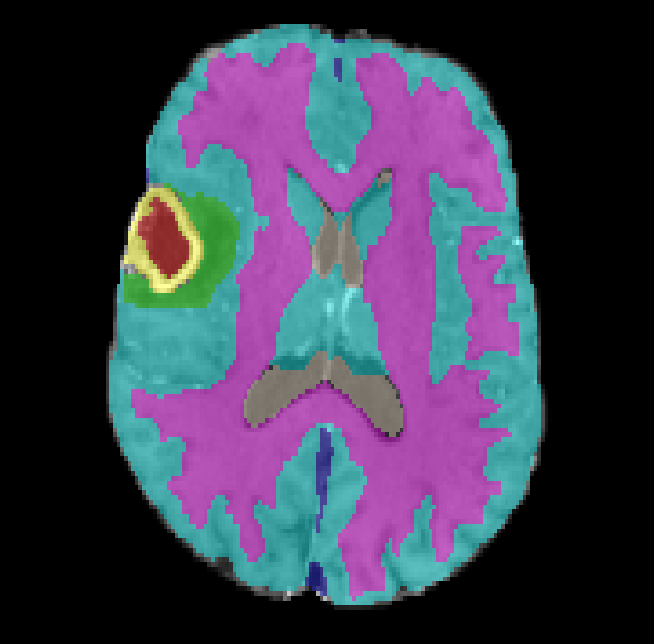}
\includegraphics[width=0.3\textwidth]{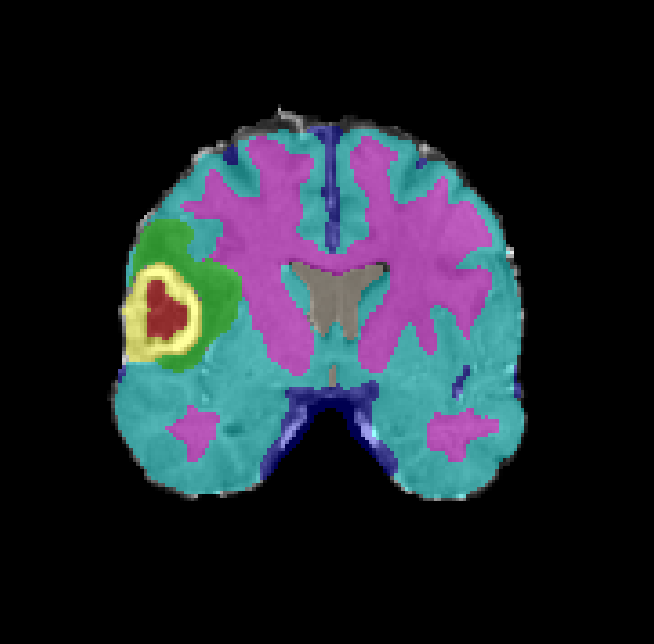}
\includegraphics[width=0.3\textwidth]{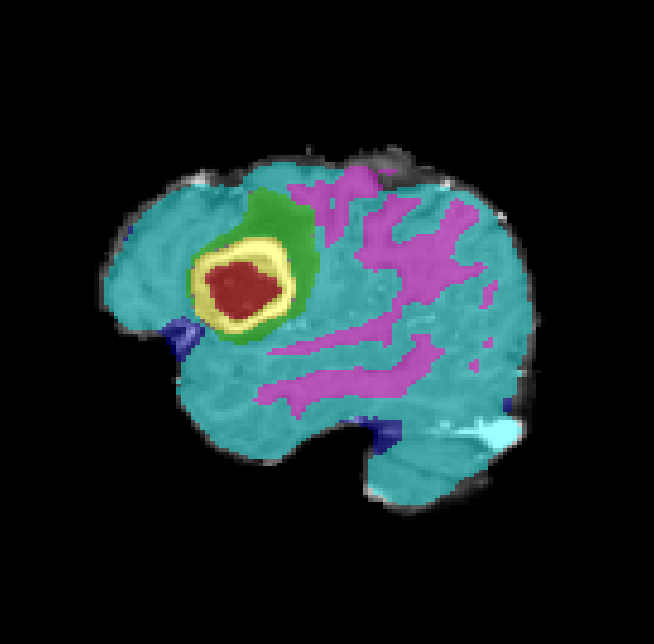}
\caption{
  (\textit{Top row}): The original T1ce validation MR-image for Brats18\_CBICA\_ABT\_1 data is shown for different views (axial, coronal and sagittal).
  (\textit{Bottom row}): The corresponding segmentation result for healthy cells computed by the neural network.}
\label{f:val_result}
\end{figure}

An orthogonal approach that we propose for data augmentation, is an extended segmentation BraTS  training data. That is, we segment the healthy parenchyma int gray/white matter, cerebrospinal fluid, and glial cells. The delineation of these
healthy tissues contain important information which is actually used by radiologists. 
For example, the  delineation of the tissues could be compressed due to tumor growth in  the confined space of the brain. Providing this information to
the classifier can help in better segmenting tumorous regions.
However, such data is not readily available in the BraTS  training dataset, since labelling the tumorous regions itself is laborious, let alone annotating
full healthy tissues
which is orders of magnitude more time consuming.
We propose a novel automated approach to compute this information through image 
registration.  In our method, we only need one (or preferably a few) fully segmented 
brains. Then given an input 3D brain, we perform the following automatic steps to obtain the extended segmentation:

\begin{figure}[!htbp]
\centering
\includegraphics[width=.8\textwidth]{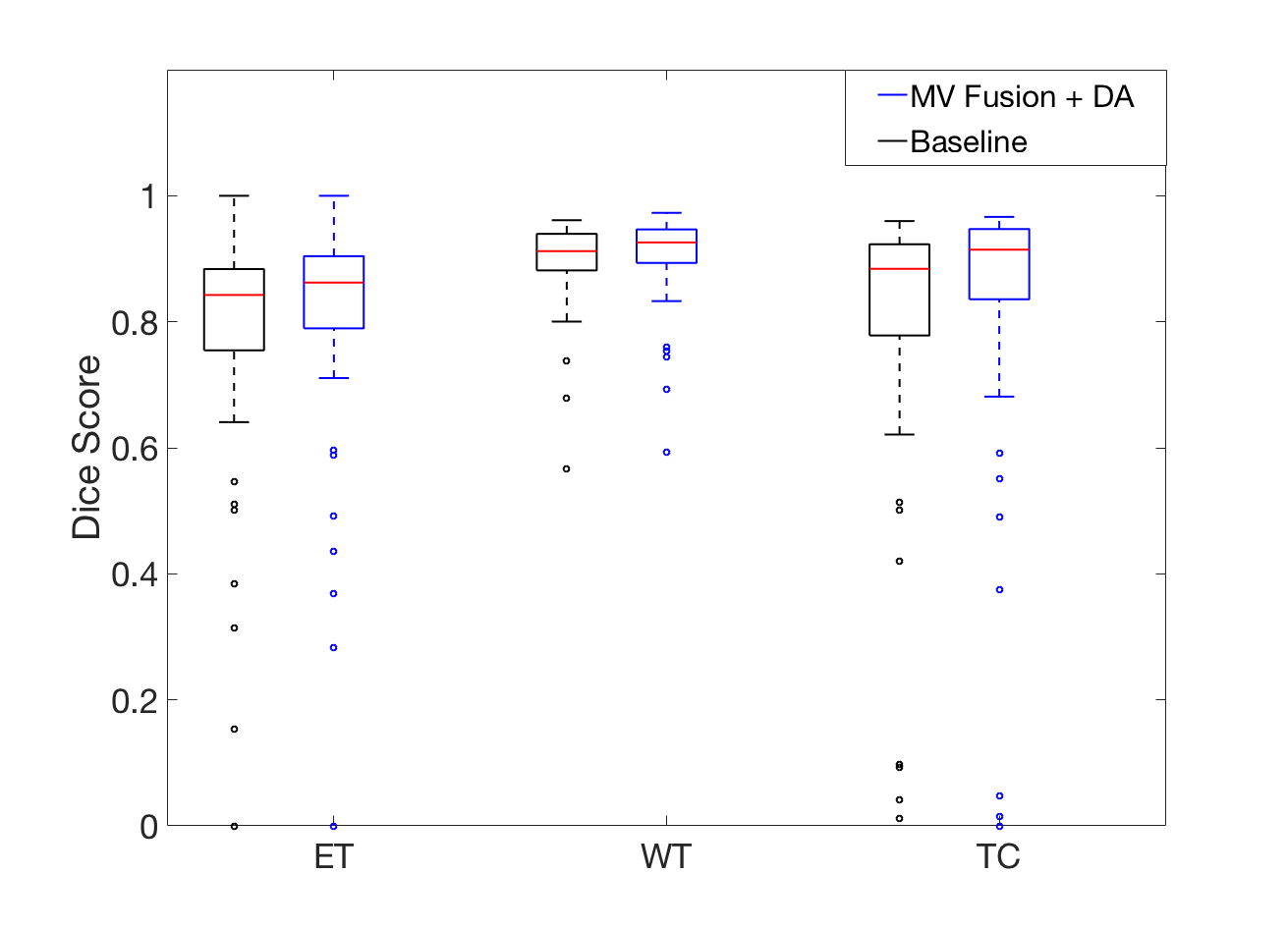}
\caption{
  Box plot for the final model's dice score on the BraTS'18 validation data is shown. This model achieves a mean dice score of (79.15,90.81,81.91) percent for (WT,TC,EN), respectively.
}
\label{f:boxplots}
\end{figure}

\begin{enumerate}   
    \item Affine registration of each atlas image to the brats image.
    \item Diffeomorphic registration of each atlas image to the BraTS image: This step aims to find a deformation map that would ``translate'' a healthy atlas to match the structure of a given BraTS training example. We compute this deformation by solving a PDE-constrained optimization problem. We refer to \cite{mang_gholami_biros_2016,mang_biros_2017,mang2018claire} for details on solving this optimization problem.
    \item Majority voting to fuse labels of all deformed atlases to get the final healthy tissue segmentation: The votes are weighted with the quality of diffeomorphic registration measured by the $L_2$ norm of the residual between each deformed atlas and brats image. This ensures the highest weight for the deformed atlas closest to the BraTS image.
\end{enumerate}

We show an exemplary segmentation for an MRI scan from the BraTS training data in Fig.~\ref{f:registration}.

\section{Results} \label{s:results}

\begin{table}[!htbp]
\footnotesize
\caption{
We report the BraTS'18 results for our method for both the baseline model and the final 2D network. Our final submission to the validation portal is highlighted. The last row shows the dice scores for BRaTS'18 testing dataset. Even though we use a sub-optimal
2D network, but we can still achieve significant
improvement with the proposed framework.
}
\begin{center}
\begin{tabular}{l | c c c c c }
                      & \multicolumn{3}{c}{{Dice Score}} \\	 
                        &        EN&WT&TC      \\	 
\midrule
\Ga Baseline    (Validation)    & 73.86&89.49&79.94\\ 
\Gd Proposed (Validation)   & 79.15&90.81&81.91\\
\midrule
\Ga Proposed (Testing) & 70.96&87.11&76.87 \\
\bottomrule
\end{tabular}
\label{t:val_result}
\end{center}
\end{table}

Here, we report our segmentation results on BRATS'18 dataset.
\subsubsection*{Baseline network for healthy and tumor segmentation}

We first obtain the healthy tissue segmentation for all the BraTS training data using 
the image registration method discussed above, and use the fine grained data to train a
neural network.  Given that our current domain adaptation framework only supports 2D 
transformations, we follow a two stage
segmentation routine using both a 3D and a 2D UNet.  The 3D U-Net has ten layers with 
multiclass dice loss (based on the works of~\cite{isensee2017}, implemented in 
TensorFlow/Keras) as the baseline network to localize the tumor. Then we train a second 
2D U-Net with the domain adaptation results to obtain the final segmentation.
The training is performed
for 500 epochs using a five-fold validation split of the training data with ADAM 
optimizer and ensemble the splits to obtain the baseline results. We show the healthy 
segmentation for a validation MRI scan in Fig.~\ref{f:val_result}.

\subsubsection*{Data augmentation through domain adaptation}
In order to avoid noisy segmentations and reduce the class imbalance inherent in the 
BraTS dataset, we create crops of the tumor regions in every slice and use them for 
training. During testing, we use the segmentation generated by our baseline 3D model to 
create crops around the whole tumor. We use fixed sizes for our crops (specifically 
$48\times48, 96\times96$ and $144\times144$). This is to ensure no loss of information 
due to strided operations when we go deeper in the neural network. We train three U-Nets
corresponding to the axial, sagittal or coronal view of the MRI scan and ensemble them 
(similar to the multi view fusion method outlined in~\cite{Wang2017}). As before, we 
train five-fold cross-validation splits and ensemble them to avoid overfitting to the 
training data.

To augment our data with domain adaptation results, we simulate a
synthetic tumor in our atlas corresponding to the whole tumor center of mass of every BraTS training image. We transfer the synthetic 
brain to the BraTS domain 
for every axial slice. Hence, our augmented dataset consists of approximately twice the 
amount of training brains. 
Our final neural network is the 2D multiview (MV) U-Net with masks generated using the 
baseline and data augmentation 
using domain adaptation.

\subsubsection*{Results}
\begin{figure}[!htbp]
\centering
\includegraphics[width=0.3\textwidth]{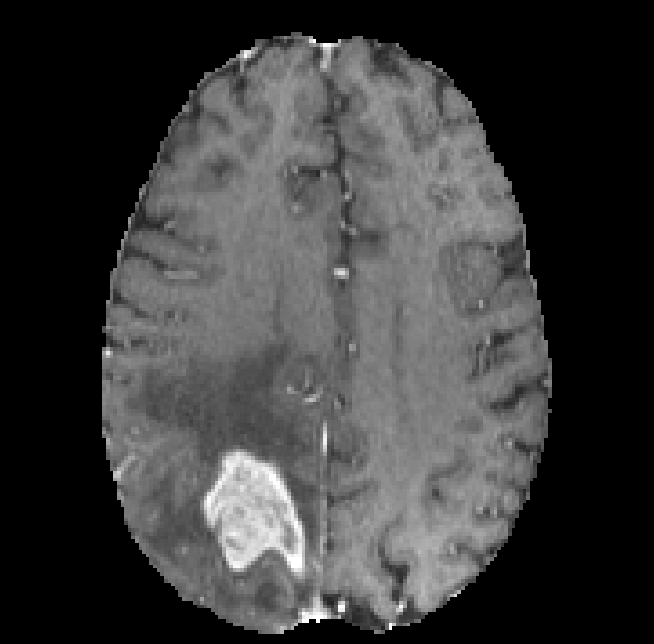}
\includegraphics[width=0.3\textwidth]{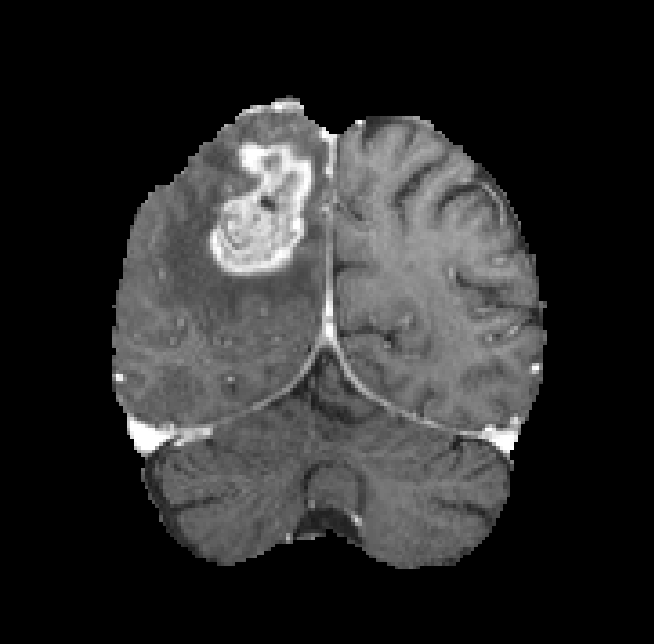}
\includegraphics[width=0.3\textwidth]{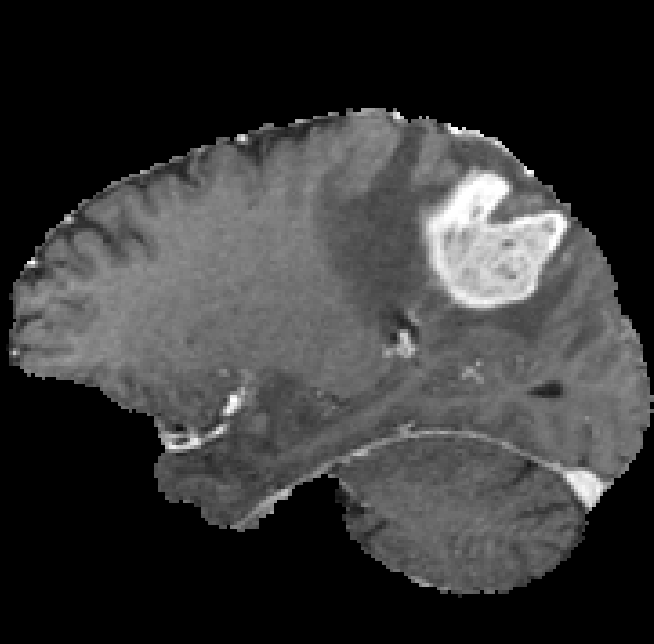}
\includegraphics[width=0.3\textwidth]{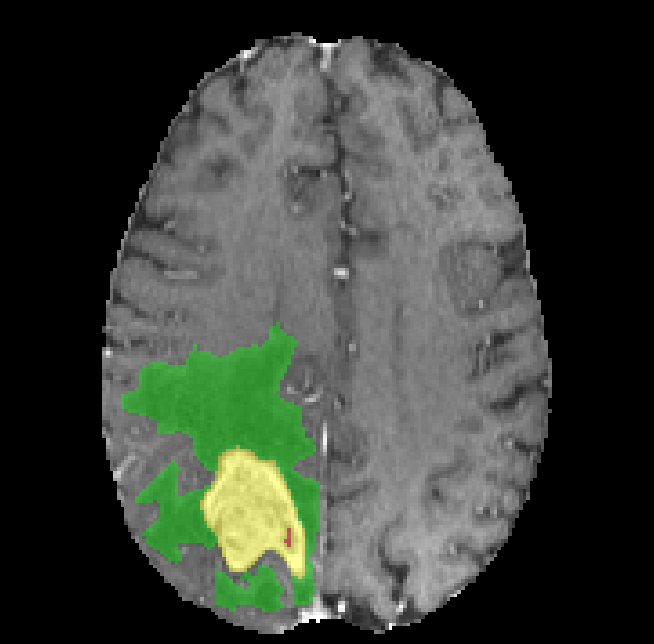}
\includegraphics[width=0.3\textwidth]{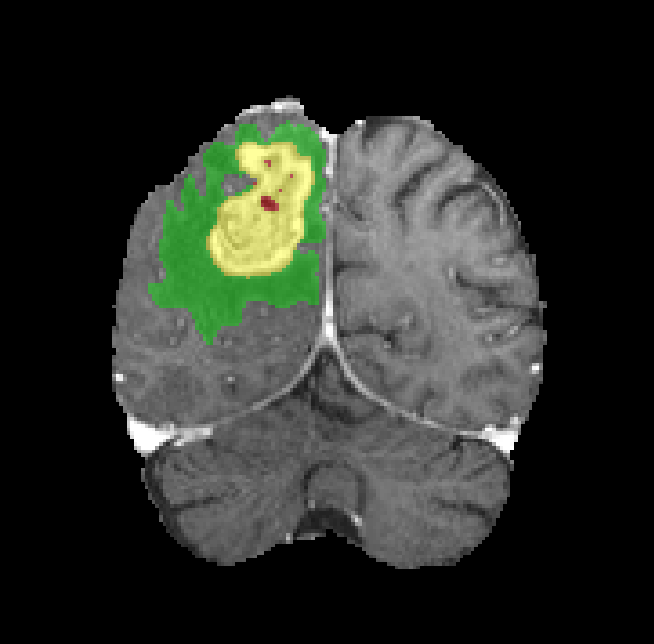}
\includegraphics[width=0.3\textwidth]{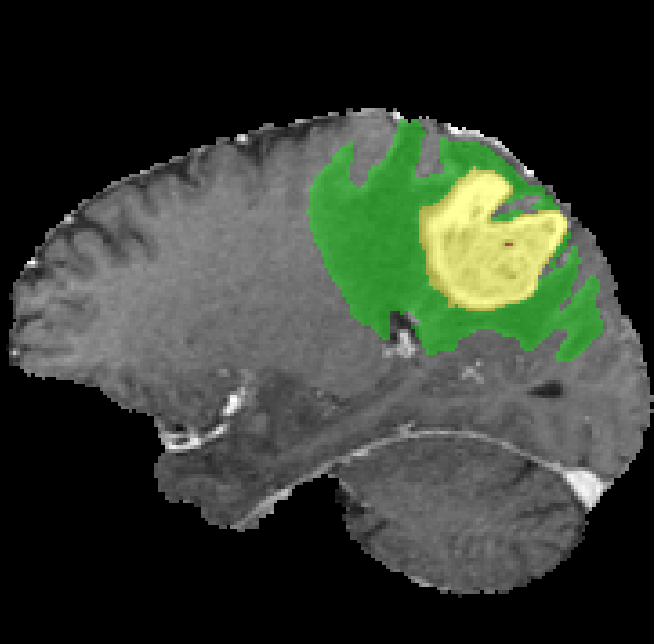}
\caption{
  (\textit{Top row}): The original T1ce validation MR-image for Brats18\_CBICA\_AAM\_1 data is shown for different views (axial, coronal and sagittal).
  (\textit{Bottom row}): The corresponding tumor segmentation result from the final 2D network.}
\label{f:val_result_2D}
\end{figure}

We trained the framework using the BraTS'18 data. The fine-grained segmentation result 
from the first stage 3D U-Net is shown in Fig.~\ref{f:val_result}. As one can see, this 
involves both the tumor segmentation, shown in red/yellow/green, as well as healthy 
structure of the brain shown in purple/cyan/gray. This data is used for localizing the 
tumor boundaries. We then use this data and create multiview slices around the tumor 
bearing region. The multiviews include the three directions of axial, sagittal, and 
coronal  directions. Then, this data is passed through the second stage 2D U-Net which 
was trained along with the domain adaptation data, and
fused together to obtain the final segmentation as shown in Fig.~\ref{f:val_result_2D}.

We show quantitative values for the Dice score in Tab.~\ref{t:val_result}, with the 
corresponding box plots shown
in Fig.~\ref{f:boxplots}. The baseline network
has a dice score of [73.86,89.49,79.94] for Enhancing Tumor (ET), Whole Tumor (WT), and 
Tumor Core (TC). Using our proposed data augmentation framework leads to a dice score of
[79.15,90.81,81.91].  These could be further improved by 
using a 3D network instead of a 2D one, by developing a 3D domain adaptation framework which is part of our
future work.

\section{Conclusion} \label{s:conclusion}

We presented a new framework for biophysics-based medical image segmentation.
Our contributions include an automatic healthy tissue segmentation of the BraTS dataset, and a novel Generative
Adversarial Network to enrich the training dataset using a model to generate synthetic phenomenological structures of
a glioma. We demonstrated that our approach yields promising results on the BraTS'18 validation dataset. Our 
framework is not specific to a particular model, and could be used with other proposed neural networks for the BraTS 
challenge.

\bibliographystyle{splncs03}
\bibliography{ref}

\end{document}